\documentclass[11pt]{article}

\usepackage[final]{acl}

\usepackage{times}
\usepackage{latexsym}

\usepackage[T1]{fontenc}
\usepackage{textcomp}

\usepackage[utf8]{inputenc}

\usepackage{microtype}

\usepackage{inconsolata}

\usepackage{graphicx}

\usepackage{amsmath}
\usepackage{stfloats}
\usepackage{booktabs}
\usepackage{tikz}
\usetikzlibrary{positioning, arrows.meta}
\usepackage{adjustbox}
\usepackage{hyperref}

\title{SynFlow: A Multidimensional Diachronic Semantic Analysis Toolkit}

\author{
  \textbf{Bach Phan-Tat}\textsuperscript{1},
  \textbf{Kris Heylen}\textsuperscript{1,2},
  \textbf{Dirk Geeraerts}\textsuperscript{1},
  \textbf{Stefano De Pascale}\textsuperscript{1,3},
  \textbf{Dirk Speelman}\textsuperscript{1}
  \\[1ex]
  \textsuperscript{1}Department of Linguistics, KU Leuven\\
  \textsuperscript{2}Instituut voor de Nederlandse Taal\\
  \textsuperscript{3}Vrije Universiteit Brussel\\[1ex]
  \small \textbf{Correspondence:} \href{mailto:ttbach.phan@kuleuven.be}{ttbach.phan@kuleuven.be}
}

\begin{document}
\maketitle
\begin{abstract}
Lexical semantic change (LSC) is commonly modelled through vector-space representations, but these approaches often provide limited insight into which aspects of usage are changing. Diachronic corpus research instead examines interpretable dimensions such as syntactic behaviour, morphology, and constructional patterns, but typically through separate analytical workflows. We present \textbf{SynFlow}, an open-source toolkit for multidimensional diachronic analysis of linguistic usage. SynFlow represents linguistically enriched word concordances as period-specific distributions and applies a shared workflow across dependency-based co-occurrences, morphological features, constructional configurations, and semantic patterns such as those from Frame Semantics. It supports different distance measures, together with value-level contribution scores, statistical testing, and incremental clustering of lexical slot-fillers. We demonstrate SynFlow through a qualitative case study of the German adjective \textit{viral}, showing how a single semantic development is reflected across syntactic, lexical, constructional, and morphological dimensions. We further report previously published results on SemEval-2020 Task 1 to situate the performance of these representations relative to existing lexical semantic change detection systems.

\end{abstract}

\section{Introduction}
Recent computational work on lexical semantic change (LSC) is dominated by vector-space modelling approaches that compare vector representations of target words across historical periods \cite{periti_lexical_2024, tahmasebi_computational_2021, tahmasebi_computational_2023}. While effective on shared tasks, such methods are difficult to interpret \cite{lenci_comparative_2022}, making their outputs difficult to relate directly to linguistic theories of semantic change: a model may indicate that a word has changed without directly revealing which aspects of its usage account for that change.

We introduce SynFlow, an open-source toolkit for multidimensional diachronic semantic analysis. SynFlow represents each dimension as a period-specific distribution of observable values and quantifies semantic change as differences between the diachronic distributions within a dimension. It natively supports automatic extraction and quantification of dependency-based co-occurrences, morphological features, and constructional configurations, while other semantic patterns such as those from Frame Semantics can be incorporated through the same interface. This allows different dimensions to be analysed within a common diachronic workflow rather than requiring a separate analytical procedure for each representation.

SynFlow compares these distributions across periods using cosine distance, Jensen--Shannon divergence (JSD) \cite{menendez_jensen-shannon_1997}, or total variation distance (TVD), and decomposes change scores into contributions from individual values, making it possible to determine \textbf{which dimensions change} and \textbf{what drives the change}. For lexical fillers, it additionally provides incremental clustering to trace broader thematic developments. The system therefore preserves a direct connection between quantitative change signals and the linguistic evidence underlying them.

The main contribution of SynFlow is a \textbf{shared workflow for multidimensional diachronic semantic analysis}, providing temporal comparison, statistical testing, and value-level contribution scores.

We demonstrate \footnote{\url{https://www.youtube.com/watch?v=CRlG2kgdfTE}} SynFlow with a qualitative case study of the German adjective \textit{viral}, showing how one semantic development is reflected across multiple linguistic dimensions. Previously published SemEval-2020 Task 1 results \cite{tat-etal-2026-transparent, tat-etal-2026-reframe} are also reported to situate the performance of our system relative to other existing LSC systems.

\section{Related work}
\subsection{Linguistic dimensions in diachronic analysis}

Theory-driven studies of semantic change have long examined linguistically motivated dimensions such as prototype structure, metaphor, frame semantics and constructions \citep[e.g.,][]{geeraerts_diachronic_1997, jansegers_towards_2020, law_diachronic_2019, petterssontraba2022development, sovran_polysemy_2004, sullivan_metaphoric_2007, sweetser_etymology_1990, vais_diachronic_2022}. While these approaches provide detailed accounts of semantic development, their analytical dimensions are often tailored to particular lexical items or change phenomena, limiting systematic comparison across targets and corpora. Computational work has begun to operationalise structured linguistic information at a larger scale \citep[e.g.,][]{kutuzov_grammatical_2021}. SynFlow extends this direction by providing a reusable workflow for comparing the temporal distributions of such structured usage dimensions within a common framework.

\subsection{Computational tools for linguistic profiling}
SynFlow is related to corpus tools that profile lexical behaviour, including Sketch Engine's Word Sketch Difference (WSD) \cite{word_sketch_difference_2019}, Korp \cite{borin_korp_2025}, and DiaCollo \cite{jurish_diacollo_2015}. DiaCollo models diachronic change through linear-window collocations which often suffer from accidental co-occurrences \cite{evert_corpora_2008}. WSD and Korp use association measures that rank individual collocates whereas SynFlow quantifies redistribution of \textbf{complete profiles}, and decomposes change into individual contributions. SynFlow also processes multiple temporal periods within \textbf{a single pass} rather than requiring repeated pairwise comparisons like WSD.

There are other tools operating on structured linguistic information in parsed corpora. ANNIS \cite{chiarcos_annis_2009}, Grew-match \cite{guillaume_graph_2021}, and the FrameNet Brasil WebTool \cite{torrent_flexible_2025} support querying, visualisation, or annotation; STARK \cite{krsnik-dobrovoljc-2025-stark} extracts configurable dependency patterns; Profiling-UD \cite{brunato-etal-2020-profiling} derives multi-level linguistic profiles; ComparaTree \cite{tercon-dobrovoljc-2025-comparatree} compares linguistic properties across treebanks; and DELTA \cite{esteve-dobrovoljc-2026-delta} combines dependency-pattern extraction with multi-level diversity measurement across corpora. However, these systems primarily support querying, extraction, profiling, annotation, or corpus-level comparison rather than \textbf{target-oriented diachronic semantic analysis}. SynFlow complements them by utilising structured linguistic information to model diachronic semantics. It models linguistic observations as target-conditioned temporal distributions and analyses their redistribution through statistical testing and value-level contribution scores.

\section{System architecture}
\begin{figure*}[t]
\centering

\begin{adjustbox}{max width=0.95\textwidth}
\begin{tikzpicture}[
    node distance=0.7cm,
    box/.style={
        draw,
        rounded corners=2pt,
        minimum height=0.9cm,
        text width=2.5cm,
        align=center,
        font=\small
    },
    arrow/.style={
        -{Stealth[length=2.5mm]},
        thick
    }
]

\node[box] (query) {Query\\(e.g., viral/ADJ)};

\node[box, right=of query]
    (dimension) {Dimension\\Extraction\\(Output: dataframe)};

\node[box, right=of dimension]
    (distribution) {Distribution\\Construction};

\node[box, right=of distribution]
    (detection) {Change\\Detection\\(cosine distance, JSD, TVD)};

\node[box, right=of detection]
    (decomposition) {Change\\Decomposition\\(item-level contributions)};

\draw[arrow] (query) -- (dimension);
\draw[arrow] (dimension) -- (distribution);
\draw[arrow] (distribution) -- (detection);
\draw[arrow] (detection) -- (decomposition);

\end{tikzpicture}
\end{adjustbox}

\caption{Overview of the workflow.}
\label{fig:pipeline}

\end{figure*}
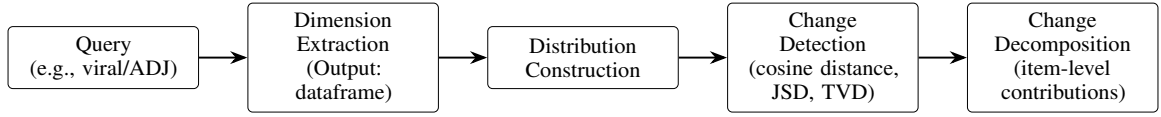

SynFlow takes annotated diachronic subcorpora as input, with each directory representing one temporal period. Given a target lemma, POS, and selected dimension types, it extracts the corresponding linguistic patterns, converts them into period-specific distributions, measures change across periods, and decomposes the resulting scores into contributions from individual values. The overall workflow is demonstrated in figure \ref{fig:pipeline}.

SynFlow distinguishes between a \textit{dimension type}, which specifies the linguistic information to extract, and an individual \textit{dimension}, whose observed values form a distribution. For target $w$, dimension $d$, and period $t$, usage is represented as $P_t(v \mid d,w)$ over values $v$. Some dimension types such as \texttt{construction} define a single distribution, while others define multiple dimensions, such as those conditioned on a specific dependency slot or morphological feature type.

SynFlow natively extracts dependency-based, constructional, and morphological information, while externally annotated dimensions can be analysed through the same downstream workflow provided their dataframe format matches the output of the Dimension Extraction step. These dimensions provide complementary views of linguistic usage. Slot and feature types describe broader structural profiles, slot fillers and morphological features capture variation within particular categories, and constructional configurations capture how multiple syntactic relations combine within the same usage instance. Dedicated notebook templates are provided for the supported analysis types.

\subsection{Dimension extraction}
\paragraph{Input format}
\label{sec:input_format}
SynFlow uses a CoNLL-U-like format in which each sub-corpus is divided into individual sentences. Each sentence begins with <s=sentence\_id> and ends with </s>. Seven tab-separated columns are required in the following order: \textsc{token, lemma, pos, id, head\_id, deprel, feats}. We have also provided a notebook and scripts for parsing the raw data (1 sentence/line) into the correct SynFlow format using Stanza \cite{qi-etal-2020-stanza}.

\begin{table*}[t]
\centering
\begin{tabular}{llll}
\toprule
\textbf{Type} & \textbf{Dimension} & \textbf{Example values} & \textbf{Interpretation} \\
\midrule
Slot type
& \texttt{SlotType}
& \texttt{chi\_nsubj}, \texttt{chi\_obj}
& grammatical profile \\

Slot filler
& \texttt{FILLER[chi\_obj]}
& cake
& lexical preferences \\

Construction
& \texttt{Construction}
& \texttt{[chi\_nsubj + chi\_obj]}
& multi-slot configurations \\

Feature type
& \texttt{FeatureType}
& Number, Tense
& morphological profile \\

Feature
& \texttt{FEAT[Number]}
& Sing, Plur
& variation within feature \\
\bottomrule
\end{tabular}
\caption{Dimensions and example values natively extracted by SynFlow.}
\label{tab:dimensions}
\end{table*}

\paragraph{Dimensions} SynFlow natively extracts five types of linguistic information (Table~\ref{tab:dimensions}) using different tree-traversal algorithms. \textbf{Slot types} are individual dependency relations involving the target, with direction encoded as \texttt{chi\_} for children and \texttt{pa\_} for parents. For example, in \textit{the boy eats the cake}, \textit{eat} is associated with \texttt{chi\_nsubj} and \texttt{chi\_obj} as it is the head of \textit{boy} and \textit{cake}, while \textit{boy} is associated with \texttt{pa\_nsubj} and \texttt{chi\_det}. \textbf{Slot fillers} are the lexical items occupying a dependency slot, with each slot defining a separate dimension. \textbf{Constructional configurations} represent combinations of dependency slots co-occurring around the same target instance, e.g.\ \texttt{[chi\_nsubj + chi\_obj]}, within a user-specified dependency-path depth. \textbf{Morphological feature types} are categories extracted from the target's \textsc{feats} field, while \textbf{morphological features} represent the observed values of each category as separate dimensions.

\subsection{Change detection}
Counts within each dimension and period are normalised to sum to one, yielding probability distributions. This differs from occurrence-normalised frequencies used in some descriptive visualisations, where counts are divided by the number of target occurrences. SynFlow then measures distribution change between consecutive periods using cosine distance, JSD, or TVD. Because distances estimated from sparse observations can be unstable, SynFlow optionally applies support weighting:
\[
Weighted Dist = RawDist \times \min\left(1,\frac{c}{k}\right),
\]
where $c=\min(c_1,c_2)$ is the smaller number of observations in the two periods, and $k$ is a user-defined support threshold. Distances receive full weight (i.e., 1) when both periods contain at least $k$ observations and are linearly down-weighted otherwise.

SynFlow also supports permutation testing to help distinguish meaningful changes from random fluctuations. Alongside the magnitude of the detected change, it provides a $p$-value indicating how likely a change of similar or greater size would be observed by chance. This gives users an interpretable measure of whether a detected change is statistically significant.

For multi-period analyses, SynFlow applies the Benjamini--Yekutieli (BY) procedure \cite{benjamini_control_2001} to control the False Discovery Rate (FDR). For each dimension, correction is applied jointly to the $p$-values from all adjacent-period comparisons.

\subsubsection{Incremental slot filler clustering}
This section only applies to slot fillers. To distinguish lexical replacement from broader thematic change, SynFlow provides incremental clustering of slot fillers. The analysis uses static word2vec embeddings \cite{mikolov2013efficientestimationwordrepresentations} trained separately for each period and Procrustes aligned \cite{smith2017offline} to a common space, following diachronic embedding approaches \citep{hamilton_diachronic_2016}. Fillers in the first period are grouped using hierarchical agglomerative clustering, with each cluster represented by the frequency-weighted centroid of its members. In subsequent periods, fillers are assigned to sufficiently similar existing centroids, while unmatched fillers form new clusters. Active centroids are then updated from the current-period fillers. Previously established clusters are retained when temporarily inactive, allowing them to be reactivatable in later periods. This incremental memory enables SynFlow to trace the emergence, persistence, decline, and reactivation of broader filler themes over time. A dedicated notebook is provided for training and aligning the required word2vec embeddings.

\subsection{Change decomposition}
After computing the distance between two period-specific distributions, SynFlow can further decompose the observed change into contributions from individual values. This allows users to move from a single aggregate change score to the specific linguistic items responsible for that difference. For a dimension $d$, SynFlow reports how much each value $v$ contributes to the overall distance between two consecutive periods. For example, if the distribution of \texttt{FILLER[chi\_amod]} changes substantially between two periods, decomposition can reveal which adjectival fillers account for most of that shift. This step is particularly useful for qualitative analysis, because it links the numerical change score back to interpretable linguistic evidence. Rather than only indicating that a dimension has changed, SynFlow shows which individual values drive the observed redistribution.

\subsection{Output and visualisation}
Most intermediate and final outputs produced by SynFlow are stored as comma-separated values (CSV) files. This allows users to inspect the extracted information directly, verify intermediate processing steps, and reuse the outputs in external analyses. The accompanying notebooks also provide visualisations at different stages of the workflow to facilitate exploration and interpretation. These include time-series plots for tracking changes across periods, heatmaps for comparing patterns across dimensions or time, and bar plots for inspecting distributions and item-level contributions. Together, the tabular outputs and visualisations allow users to move between quantitative summaries and the linguistic evidence underlying them.

\subsection{Availability and execution}
SynFlow is freely available as open-source software under the MIT license. The repository \footnote{\url{https://github.com/phantatbach/SynFlow/releases/tag/v3.0.0}}  provides installation instructions, dependency specifications, documentation of the expected corpus format, and ready-to-use notebook templates for the main analysis types. SynFlow is currently distributed as a notebook-based Python workflow. Users install the system by cloning the repository, creating a Python environment, installing the packages specified in \texttt{requirements.txt}, and running the notebooks. Python 3.10 or newer is recommended.

There are dedicated notebooks \footnote{The dataset and type embeddings are too large to upload to public servers. We have provided dataframes of different dimensions so that users can explore Change Detection and Change Decomposition.} for \texttt{slot-level}, \texttt{constructional}, \texttt{morphological-feature}, \texttt{diachronic embedding} and \texttt{qualitative analysis}. Within each notebook, users specify the corpus location, target lemma and POS, temporal periods, and relevant analysis parameters. Input corpora are organised into directories corresponding to temporal periods and follow the seven-column CoNLL-U-like format described in \ref{sec:input_format}. The notebooks then guide users through the complete workflow from dimension extraction and distribution construction to change measurement, decomposition, statistical analysis, and visualisation. This design allows the different linguistic dimensions to be analysed through a common workflow without requiring users to modify SynFlow's underlying extraction and analysis functions.

\subsection{Runtime and scalability}
SynFlow supports multiprocessing at every possible step. As a practical benchmark, we ran SynFlow with a representative target (plane/NOUN) from Corpus of Historical American English \cite{davies2010coha} (1810s - 2000s, divided into 20 periods and containing 116,613 parsed files), on a personal laptop (32 GB RAM, AMD Ryzen 7 8845HS). Direct slot-level exploration, which reads the parsed corpus, counts dependency slots within each period, and produces the corresponding visualisations, took approximately 7 minutes 30 seconds. Extracting all slot fillers and constructing the slot-filler dataframe took 4 minutes 55 seconds. For each adjacent-period transition, we ran 1,000 permutations for each pair of slot-filler distributions of every extracted slot. Processing all slot-specific permutation tests for one transition required only 3 minutes. These results indicate that SynFlow remains practical for large diachronic corpora on consumer hardware.

\section{Evaluation}
\subsection{Demonstrative case study}

We demonstrate SynFlow with the German adjective \textit{viral} in the Leipzig German News corpus \cite{wortschatz_leipzig_downloads_2026}, covering 1995--2025 and divided into six five-year periods. Over this period, \textit{viral} shifts from predominantly virus-related uses, as in \textit{virale Infektion} ‘viral infection’, towards the newer sense ‘widely circulated or popular’, especially in online and social-media contexts, as in \textit{viral gehen} ‘go viral’.

\begin{figure}[htbp]
    \centering
    \includegraphics[width=\columnwidth]{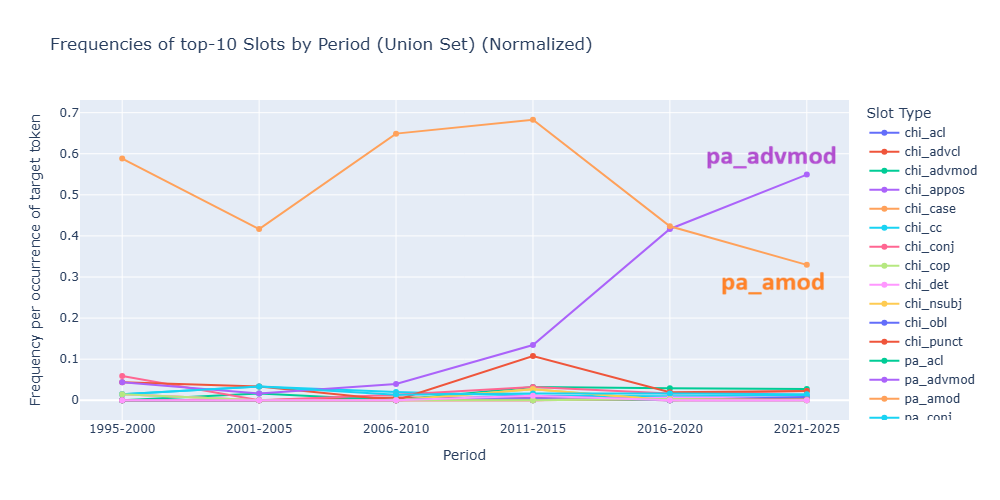}
    \caption{Change of normalised frequencies of all slot types of \textit{viral}.}
    \label{fig:slot_types_diachronic_freq}
\end{figure}

In the \texttt{slot type} dimension, occurrence-normalised frequencies of different slots (Figure~\ref{fig:slot_types_diachronic_freq}) remain largely stable except for a marked shift around 2011--2015: \texttt{pa\_amod} decreases while \texttt{pa\_advmod} increases, reflecting the growing use of \textit{viral} as an adverbial modifier in expressions such as \textit{viral gehen} and a relative decrease of the use as an adjectival modifier like in \textit{virale Infektion}.

\begin{figure}[htbp]
    \centering
    \includegraphics[width=\columnwidth]{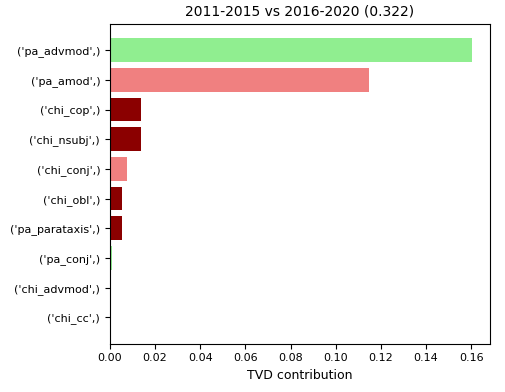}
    \caption{Contributions of individual slot types to TVD of \textit{viral}. Green bars indicate an increase in relative frequency; red bars indicate a decrease.}
    \label{fig:slot_type_contrib}
\end{figure}

This redistribution produces a large TVD between the two periods, and decomposition shows that the redistribution of \texttt{pa\_amod} and \texttt{pa\_advmod} accounts for much of the overall TVD (Figure~\ref{fig:slot_type_contrib}).

\begin{figure}[htbp]
    \centering
    \includegraphics[width=\columnwidth]{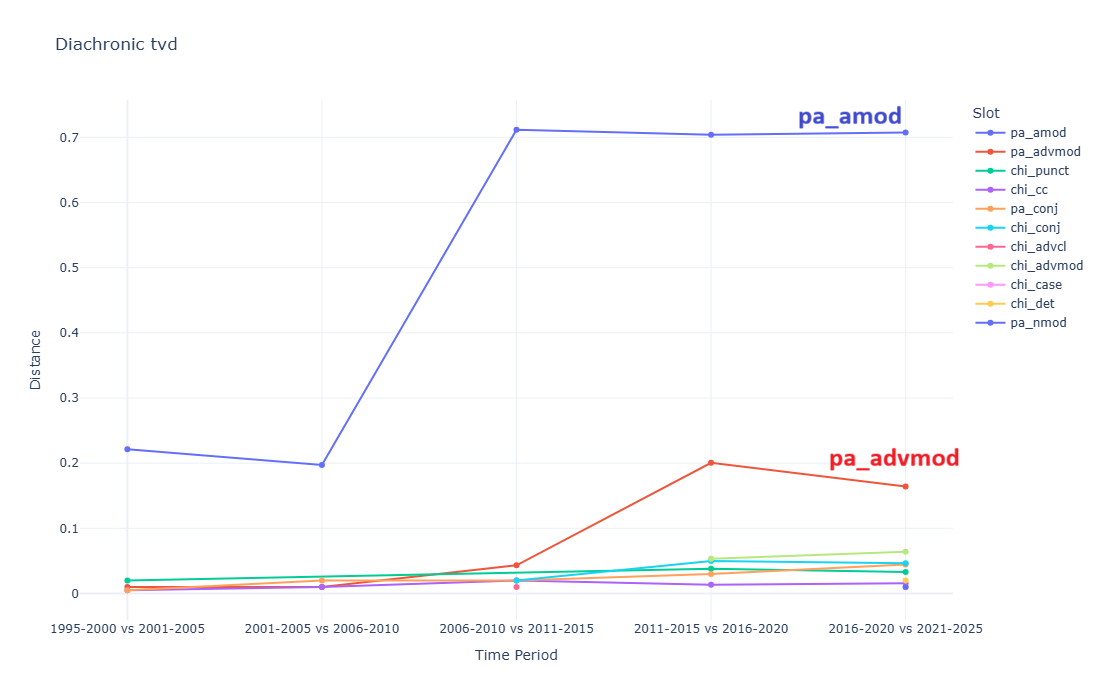}
    \caption{Weighted TVD trajectories for slot-filler distributions across dependency slots.}
    \label{fig:diachronic_TVD_all_slots}
\end{figure}

\begin{figure}[htbp]
    \centering
    \includegraphics[width=\columnwidth]{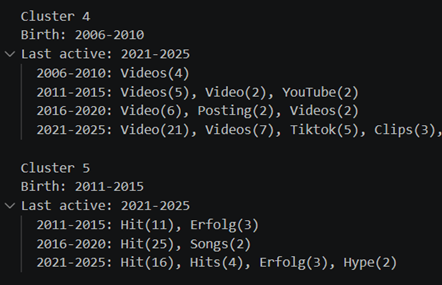}
    \caption{Incremental clustering results with social-media-related clusters growing rapidly from 2011.}
    \label{fig:incremental_clustering.png}
\end{figure}

The \texttt{slot-filler} dimensions of \texttt{pa\_amod} and \texttt{pa\_advmod} show the largest weighted TVD increases around 2011--2015 (Figure~\ref{fig:diachronic_TVD_all_slots}), while most other slots remain comparatively stable. Further decomposition and clustering analysis (Figure \ref{fig:incremental_clustering.png}) reveal that these changes are increasingly associated with fillers from online and social-media discourse, including \textit{Hit, Marketing, Video, TikTok} in \texttt{pa\_amod}, and \textit{gehen} in \texttt{pa\_advmod}. Slot-filler analysis therefore complements the structural signal by showing the lexical contexts accompanying the newer meaning of \textit{viral}.

\begin{figure*}[t] 
  \centering 
  \includegraphics[width=0.7\textwidth]{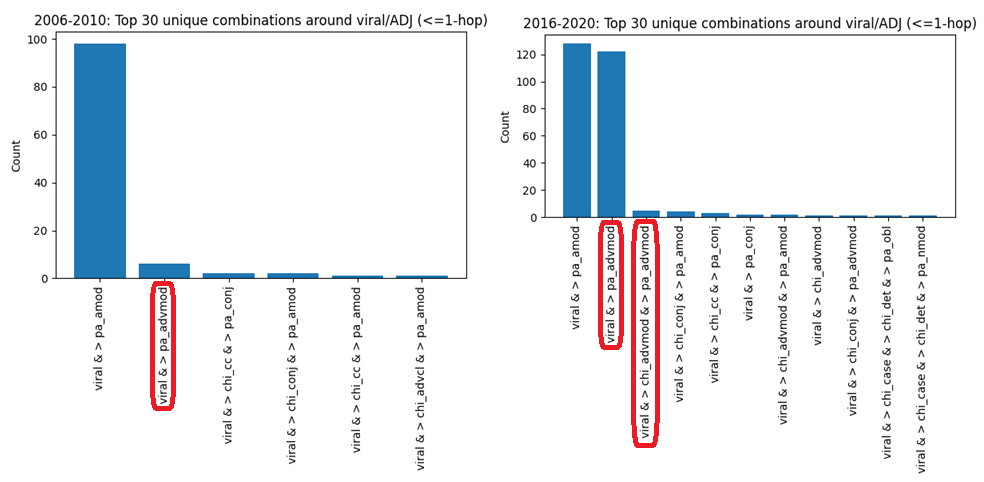} 
  \caption{Frequencies of different constructions (combinations of directly dependent slots) around \textit{viral}} 
  \label{fig:construction_freq_comparison} 
\end{figure*}

The same development is visible at the \texttt{constructional} level, showing that the rise of the adverbial \textit{viral} is not limited to an isolated dependency relation, but also affects the larger combinations of grammatical relations in which the target occurs. From around the same period, configurations containing \texttt{pa\_advmod} become increasingly frequent (Figure \ref{fig:construction_freq_comparison}). For example, the configuration \texttt{[chi\_advmod + pa\_advmod]} occurs in expressions such as \textit{gerade viral gehen} ‘to be going viral right now’, where \textit{viral} is linked to \textit{gehen} through \texttt{pa\_advmod} and to \textit{gerade} through \texttt{chi\_advmod}.

At the \texttt{morphological} level, the frequency of \texttt{Variant} increases around the same period as \texttt{pa\_advmod}. These adverbial uses are annotated as \texttt{Variant=Short} as they lack the normal adjectival inflection because they are used as an English loan. However, \texttt{FEAT[Variant]} contains only the value \texttt{Short}, meaning that the change is visible at the feature-type level, but not in the corresponding feature-value dimension. This contrasts with the dependency dimensions, where change is visible both at the broader structural level and within individual slot-filler distributions.

Overall, the case of \textit{viral} illustrates how one semantic development can leave complementary traces across several linguistic dimensions. Rather than collapsing these signals into a single score, SynFlow allows users to identify where change occurs and inspect the dimension-specific evidence.

\subsection{Experimental benchmark}
We previously benchmarked SynFlow on SemEval-2020 Task 1 \cite{schlechtweg_semeval-2020_2020} using two linguistic representations: dependency-based slot fillers and Frame Semantics \cite{tat-etal-2026-transparent, tat-etal-2026-reframe}. Both representations proved effective for lexical semantic change detection and ranked above many neural-network-based systems on the benchmark (Table \ref{tab:experimental_benchmark}), while retaining direct interpretability of the linguistic evidence underlying the change scores.

\begin{table}[t]
\centering
\begin{tabular}{lcc}
\hline
\textbf{Representation} & \textbf{Subtask 1} & \textbf{Subtask 2} \\
\hline
Slot filler & 4th /28 & 12th /28 \\
Frame Semantics & 9th /28 & 9th /28 \\
\hline
\end{tabular}
\caption{Ranking of previously evaluated SynFlow representations against systems reported for the English SemEval-2020 benchmark. Full results are reported in \citet{tat-etal-2026-transparent, tat-etal-2026-reframe}.}
\label{tab:experimental_benchmark}
\end{table}

\section{Conclusion and future work}
We presented SynFlow, an open-source toolkit for multidimensional diachronic semantic analysis. It natively supports dependency-based, constructional, and morphological information while allowing other semantic patterns to be analysed through the same interface. The \textit{viral} case study illustrates how a single semantic development can be traced across multiple dimensions, while previous benchmark results show competitive performance against embedding-based methods. Future work will extend SynFlow to additional linguistic dimensions and integrate further parsers and LLM-based annotation methods into the extraction pipeline.


\bibliography{custom}




\end{document}